\documentclass[11pt,letterpaper]{article}
\usepackage{cogsys}
\usepackage[T1]{fontenc}
\usepackage{times}
\usepackage[pdftex]{graphicx} 

\usepackage{xcolor}
\usepackage[dvipsnames]{xcolor}
\usepackage{listings}
\lstdefinestyle{mystyle}{
    commentstyle=\color{OliveGreen},
    keywordstyle=\color{Magenta},
    numberstyle=\tiny\color{Gray},
    stringstyle=\color{Purple},
    basicstyle=\fontfamily{lmtt}\scriptsize,
    breakatwhitespace=false,         
    breaklines=true,                 
    captionpos=b,                    
    keepspaces=true,                 
    numbers=left,                    
    numbersep=5pt,                  
    showspaces=false,                
    showstringspaces=false,
    showtabs=false,                  
    tabsize=2
}
\usepackage{natbib}
\usepackage{bm}
\usepackage{multirow}
\usepackage{amsmath}
\usepackage{amssymb}

\cogsysheading{X}{20XX}{1-6}{X/20XX}{X/20XX}

\ShortHeadings{LLM-Symbolic Continuous Causal Text Interpretation}
              {X.\ Lian and K.\ Forbus}

\begin{document} 

\title{LLM-Augmented Symbolic NLU System for More Reliable Continuous Causal Statement Interpretation}
\author{Xin Lian}{x.lian@u.northwestern.edu}
\author{Kenneth D. Forbus}{forbus@northwestern.edu}
\address{Department of Computer Science, McCormick School of Engineering and Applied Science, \\Northwestern University, Evanston, IL 60208}
\vskip 0.2in
 
\begin{abstract}
Despite the broad applicability of large language models (LLMs), their reliance on probabilistic inference makes them vulnerable to errors such as hallucination in generated facts and inconsistent output structure in natural language understanding (NLU) tasks. 
By contrast, symbolic NLU systems provide interpretable understanding grounded in curated lexicons, semantic resources, and syntactic \& semantic interpretation rules. They produce relational representations that can be used for accurate reasoning and planning, as well as incremental debuggable learning.  However, symbolic NLU systems tend to be more limited in coverage than LLMs and require scarce knowledge representation and linguistics skills to extend and maintain.  
This paper explores a hybrid approach that integrates the broad-coverage language processing of LLMs with the symbolic NLU capabilities of producing structured relational representations to hopefully get the best of both approaches.  We use LLMs for rephrasing and text simplification, to provide broad coverage, and as a source of information to fill in knowledge gaps more automatically.  We use symbolic NLU to produce representations that can be used for reasoning and for incremental learning.  We evaluate this approach on the task of extracting and interpreting quantities and causal laws from commonsense science texts, along with symbolic- and LLM-only pipelines. Our results suggest that our hybrid method works significantly better than the symbolic-only pipeline.

\end{abstract}

\section{Introduction}

Recent years have seen the dramatic advancement of large language models (LLMs), whose capabilities now span a wide range of tasks, including natural language processing, question answering, and even creative generation \citep{achiam2023gpt, team2023gemini, dubey2024llama, abdin2024phi, gunter2024apple}. 
Do they genuinely understand natural language, though? LLMs do not engage in formal, structured reasoning in the way humans do: while some studies have drawn connections between LLM mechanisms (e.g., attention and residual streams) and cognitive psychology constructs \citep{goyal2022inductive, Yu2023NeuronLevelKA}, LLMs fundamentally operate through probabilistic pattern matching rather than principled reasoning procedures \citep{jiang2024peek}, which causes inevitable issues like hallucination \citep{marcus2020next, huang2025survey}. Particularly, it has been shown that LLMs cannot learn certain fundamental semantic properties in language understanding, including semantic entailment and consistency as defined in formal semantics \citep{asher2023limits}. 

Natural language understanding (NLU) systems, predating the widespread adoption of LLMs, integrate multiple aspects of language processing -- syntax, semantics, and inference -- often through a parser grounded in hand-crafted rules and a formal model of linguistic structure \citep{winograd1972understanding}. More advanced symbolic, and in some cases hybrid, systems include NL-Soar \citep{rubinoff1994real, lindes2017cognitive}, Companion NLU (CNLU, \citealp{tomai2009ea}), and the Never-Ending Language Learner \citep{mitchell2018never}. These knowledge-based systems use structured, discrete, and inspectable representations, enabling interpretations that are explainable and supporting reasoning.
Despite these advantages, symbolic NLU systems face persistent challenges: natural language is complex and constantly evolving, while the coverage of their lexicons and semantic knowledge bases is inherently limited, sometimes incorrect, incomplete, outdated, or unsuitable for certain domains. This restricts their applicability and necessitates significant manual effort to expand rule sets and knowledge bases according to the system’s ontological framework. Although LLMs lack explicit reasoning in the symbolic sense, they still dominate language processing due to their ability to generate text through statistical pattern matching and their training on vast amounts of text. As such, they present a promising complementary resource: filling gaps in symbolic systems’ knowledge bases.

In this paper, we present our ongoing work on integrating LLMs into the CNLU system to expand its capabilities for language understanding. Our focus here is on the domain of comparative analysis questions, using texts from the QuaRTz dataset \citep{tafjord2019quartz}. We show how LLMs can be used to translate the diverse natural language forms of wild text into simplified language that CNLU can more easily process. We also show how LLMs can be used to extend the formal semantics that CNLU uses, expanding its lexicon. We compare this hybrid approach with pure CNLU and LLM baselines over the same texts, showing that the hybrid version outperforms the pure CNLU baseline and becomes more comparable to the pure LLM baseline. As this is our initial attempt to help CNLU expand semantics and achieve better performance in qualitative reasoning tasks with LLMs, we then analyze some possible ways to improve as future steps.


\section{Background}

\subsection{CNLU} \label{sec:CNLU}

The Companion Natural Language Understanding System (CNLU) \citep{tomai2009ea} is a rule-based semantic parser. It is grounded in the NextKB\footnote{For more details, see \url{https://www.qrg.northwestern.edu/nextkb/index.html}} knowledge base (KB), 
which uses the OpenCyc ontology, KB contents, and microtheory structure (microtheories provide a means of contextualization). NextKB includes a large open-license English lexicon, an integration of FrameNet \citep{baker1998berkeley, fillmore2001building} and Discourse Representation Theory, as well as support for spatial, qualitative, and analogical reasoning and learning.


CNLU parses texts using the Allen Trains chart parser \citep{allen1995natural}, processing tokens left-to-right and bottom-up while incrementally matching grammar rules to generate parse tree candidates. Lexical features are specified via facts in the KB and augmented with semantic translation (semtrans) facts that tie words to the ontology. A semtrans specifies the semantic information a word implies in context. For example, one semtrans for the comparative adjective {\scriptsize \tt cooler} encodes a comparison event whose associated quantity type is {\scriptsize \tt Temperature}, with a word frame type representing the level of force or exertion, and a comparative relation {\scriptsize \tt lessThan}:

\begin{lstlisting}[language=Lisp]
(fnsemtrans (TheList) Cooler-TheWord Adjective
  (and (isa :EVENT ComparisonEvent) (comparer :EVENT :SUBJECT)
       (comparee :EVENT :NOUN) (comparisonReln :EVENT lessThan)
       (comparisonQtype :EVENT Temperature)) (frame FN_Temperature)
  (bindingTemplate (TheList)) (groupPatterns (TheSet)))
\end{lstlisting}

\noindent These semtrans play a direct role in generating formal interpretations of each lexical token in a given sentence in choice sets, based on their roles computed by the parser and denoted by keywords (here {\scriptsize \tt :EVENT}, {\scriptsize \tt :SUBJECT}, and {\scriptsize \tt :NOUN}). For example, given the sentence ``When particles move more slowly, temperature is lower and an object feels cooler'',
the lexical token {\scriptsize \tt cooler} would encode a piece of formal representation as part of the semantic interpretation as follows, indicating the comparison event implied:
\begin{lstlisting}[language=Lisp]
(and (isa cool13002 ComparisonEvent) 
 (comparer cool13002 object11166) (comparee cool13002 object11166) 
 (comparisonReln cool13002 lessThan) (comparisonQtype cool13002 Temperature))
\end{lstlisting}

Selecting the appropriate semantic interpretation, or using them to generate the desired logical form in a given context, requires some means for disambiguation and further semantic analysis. As per \citep{tomai2009ea} these further semantic analyses can, via abductive reasoning, serve to also disambiguate at the same time. Two families of methods we have explored are: (1) analogical Q/A training with query cases \citep{crouse2018learning, wilson2019analogical, ribeiro2019predicting, crouse2021question, ribeiro2021combining}, which map semantic choices to desired logical forms for specific contexts; or (2) narrative functions \citep{mcfate2014using}, which construct logical forms based on interpretation rules and grammatical–semantic knowledge, sometimes with additional statistical methods \citep{ribeiro2021combining}.


\subsection{QP Theory and Linguistic Frames}

Qualitative Process (QP) Theory \citep{forbus1984qualitative} provides a formal theory for qualitative causal mathematics that includes the kinds of phenomena explored in QuaRTz problems. Specifically, {\it qualitative proportionalities} are partial information about functional relationships between quantities. For example, {\scriptsize \tt (qprop <temperature of gas> <speed of its particles>)}
captures part of the intended meaning of the earlier example. {\scriptsize \tt qprop} implies an increasing monotonic functional dependency, so here a lower speed would license the inference that the temperature is lower, all else being equal. These relationships are partial and need to be combined to construct a complete causal account to reason with. This compositionality of qualitative models is an excellent fit for the compositionality of language. Expressing complex ideas in language typically requires carving them up into pieces so they can be reassembled by the recipient. This makes qualitative modeling a natural approach to aspects of natural language semantics.

For constructing qualitative knowledge from language, \cite{kuehne2004understanding} demonstrated that QP theory constructs can be represented in a frame-based format compatible with the semantic frames of FrameNet, highlighting a natural alignment between the organization of physical knowledge in QP theory and the structure of frame semantics. Subsequent work extended this frame notion to a broader range of qualitative models and narrative functions for extracting them from language (e.g. \citealp{mcfate2014using}; \citealp{forbusunifying2020}). We illustrate by example, given the statement ``The Sun is large'', CNLU with narrative functions produces the following quantity frame, where {\scriptsize \tt -DV} means the discourse variable created during the analysis:

\begin{lstlisting}[language=Lisp]
((QuantityFrame4126 
  (isa QuantityFrame4126 QuantityFrame) (quantityEntity QuantityFrame4126 TheSun)
  (quantityEntityDV QuantityFrame4126 TheSun) (quantityType QuantityFrame4126 Volume)
  (quantityValue QuantityFrame4126 (HighAmountFn Volume))))
\end{lstlisting}


Similarly, to encode a qualitative proportionality between quantities, narrative functions construct an influence frame with slots for a constrainer (cause) quantity, a constrained (effect) quantity, and a sign indicating the direction of change. So in this case, QP frames would be the desired logical forms when trying to generate interpretations that describe the causal qualitative model.

\subsection{QuaRTz}  \label{sec:quartz}

One of the earliest natural language datasets focused on qualitative relationship reasoning was {\it QuaRel} \citep{tafjord2019quarel}, which contains 2,771 multiple-choice story questions paired with logical forms. However, these questions are limited to a predefined set of qualitative relations and vary only in narrative context and subject matter. To address this limitation, \citet{tafjord2019quartz} later introduced {\it QuaRTz}, the first open-domain dataset for reasoning about textual qualitative relationships. QuaRTz includes 3,864 crowdsourced, situated questions, each annotated with the properties being compared and the corresponding grounding fact, which is a natural language sentence conveying the qualitative knowledge required to answer the question. Unlike QuaRel, QuaRTz is not restricted to a fixed inventory of qualitative relations. For example, the questions:
\begin{quote}
    {\it \small (1) If Mona lives in a city that begins producing a greater amount of pollutants, what happens to the air quality in that city? (A) increases (B) decreases\\
    (2) As Cindy swam closer to the ocean surface, it got lighter and \_\_\_ (A) warmer (B) cooler}
\end{quote}
are supplemented with respective grounding facts that can be used in finding an answer:
\begin{quote}
    {\it \small (1) More pollutants mean poorer air quality.\\
    (2) Water density increases as salinity and pressure increase, or as temperature decreases.}
\end{quote}
A grounding fact may imply comparisons in two ways: (1) between different quantities treated as entities, whose amounts or degrees are compared (as in Question 1), or (2) between quantities associated with the same entity set (as in Question 2). Moreover, a grounding fact such as that for Question 2 can mention multiple constrainer and constrained quantities, thereby encoding several causal relations, even though not all of them are directly relevant to answering the question.

\subsection{Previous Work in Differential Qualitative Analysis}

One promising way to solve qualitative reasoning problems like the ones in QuaRel or QuaRTz is via comparative analysis \citep{weld1988theories}, which uses qualitative representations to infer the causal consequences of differences, either introduced hypothetically within a system or observed between two distinct physical systems. In particular, differential qualitative analysis (DQA) \citep{weld1988theories} applies a set of inference rules to compute relative values across system descriptions based on specified differences. \citet{klenk2005solving} further demonstrated that DQA can be generalized by employing analogical mappings to align the compared systems automatically. 
An LLM might also be used in solving such problems, but it would not be based on an explicit model of some phenomenon, whereas differential analysis with DQA is always based on an explicit qualitative causal model.

\citet{crouse2018learning} was the first to apply DQA, augmented with analogical mappings, to solve problems from the QuaRel dataset, using mappings between structured situation descriptions derived from natural language. Inspired by the original QuaRel design, which models comparison questions as involving two ``worlds'' \citep{tafjord2019quarel}, each represents one change in a particular quantity from one or more entities. Crouse constructed microtheories for each compared ``world'', embedding their qualitative representations accordingly. However, the ``two-world'' separation is ultimately a lazy abstraction -- the questions are schematically constructed around a fixed set of qualitative relationships, assuming two distinct quantities are always compared, whereas it breaks down with more complex questions which may involve more entities, or quantities compared. Ultimately, effective language understanding should be grounded in the semantic meanings of the entities within a sentence, rather than reduced to an artificial partition into two ``worlds''.

\subsection{Augmenting LLMs with Symbolic Systems}


The texts in QuaRel and QuaRTz are free-form, everyday texts, while the syntactic and lexical coverage of CNLU is much narrower. Prior evidence suggests that incorporating LLMs can help. For example, \cite{ribeiro2021combining} integrated BERT into Companion to assist with disambiguation in NLU for learning by reading from Simple English Wikipedia articles. They fine-tuned BERT on FrameNet data to serve as a frame classifier and further adapted it to assess fact plausibility by training it on examples of correct and incorrect textual statements.

Why not replace CNLU with an LLM trained to produce NextKB representations? First, modern ontologies are huge, which would require massive amounts of training data. For example, NextKB includes over 80,000 concepts and over 20,000 predicates, and its 700,000+ facts are distributed across over 1,300 microtheories. Second, ontologies (like language) evolve incrementally, which is incompatible with the batch learning method of LLMs. Finally, the inspectability and debuggability of symbolic systems are an important benefit. Thus, we have been exploring several alternatives. \cite{nakos2023using} proposed using LLMs to simplify text, since text-to-text tasks are natural for LLMs, and regularizing the language input should increase CNLU's coverage. A second use of LLMs treats them as an informant, getting information from them in language that the receiving system must turn into knowledge. For example, \cite{kirk2022improving} used prompt engineering to guide an LLM in producing text that could be read by a Soar agent under human oversight, and later reduced this oversight through a framework that combined LLMs with tree search \citep{kirk2024improving}.
Later, \citet{wray2024eliciting} proposed an LLM-augmented framework, the Cognitive Task Analyst, assigning LLMs more specific, curated roles in generating formal representations for defining problem spaces. In this paper, we adopt a similar philosophy: LLMs serve curated, well-defined roles in supporting semantic representations generation, thereby improving language understanding of the NLU system.

\section{Task Description}

In this paper, we focus on understanding the grounding facts for each QuaRTz question, namely the continuous causal statements describing causal relationships between quantities that are relevant to answering the question. More specifically, with LLM assistance, our hybrid approach aims to extract the quantities mentioned in each grounding fact, along with the influence signs (i.e., directions of change) for each quantity. For example, given the grounding fact ``When particles move more slowly, temperature is lower and an object feels cooler'',
we expect the system to produce a formal representation identifying the quantities and their corresponding influence signs, such as {\scriptsize \tt (speed -) (temperature -)}
, and such compact representations are acceptable under the assumptions outlined in Section~\ref{sec:quartz}.
Once the system can generate appropriate QP frames, either indicating an influence relation or an ordinal relation between compared entities for a given quantity, we can convert these frames into readable assertions and directly identify the quantity types and influence signs, as shown in the following. In this paper, we primarily focus on generating correct ordinal frames (which imply comparison events).

\begin{lstlisting}[language=Lisp]
;; Quantity frames for quantity type `Speed`
(QuantityFrame1
  (isa QuantityFrame1 QuantityFrame) (quantityType QuantityFrame1 Speed)
  (quantityEntity QuantityFrame1 particle1))
(QuantityFrame2
  (isa QuantityFrame2 QuantityFrame) (quantityType QuantityFrame2 Speed) 
  (quantityEntity QuantityFrame2 (GapFn :NOUN)))
;; Quantity frames for quantity type `Temperature`
(QuantityFrame3
  (isa QuantityFrame3 QuantityFrame) (quantityType QuantityFrame3 Temperature) 
  (quantityEntity QuantityFrame3 particle1))
(QuantityFrame4
  (isa QuantityFrame4 QuantityFrame) (quantityType QuantityFrame4 Temperature) 
  (quantityEntity QuantityFrame4 (GapFn :NOUN)))
;; Ordinal frames -- interpreted as comparison events
(OrdinalFrame1
  (isa OridinalFrame1 OrdinalFrame) (OrdinalReln OrdinalFrame1 lessThan)
  (quantity1 OrdinalFrame1 QuantityFrame1) (quantity2 OrdinalFrame1 QuantityFrame2))
(OrdinalFrame2
  (isa OridinalFrame2 OrdinalFrame) (OrdinalReln OrdinalFrame2 lessThan)
  (quantity1 OrdinalFrame2 QuantityFrame3) (quantity2 OrdinalFrame2 QuantityFrame4))
\end{lstlisting}

\section{Methodology}

The set of quantities that people deal with is vast, larger than CNLU's lexicon. Hence we start by describing how we dynamically expand the lexicon by using information extracted by an LLM on the QuaRTz training set. Then describe how QP frames are extracted for each grounding fact, and the pipelines we used for comparison: pure-LLM, pure-CNLU, and the hybrid approach, where CNLU is assisted with LLM in either rephrasing original grounding facts or expanding the lexicon. Finally, we explain the metrics used in the evaluation. Note that in our lexicon diagnosis, lexicon expansion, and experiments, Phi-4 \citep{abdin2024phi} was used where an LLM is needed, because it is relatively small and, in our observations, produces more stable outputs\footnote{Here, ``stable'' means fewer explanations or side notes; we found that larger LLMs tend to generate more irrelevant content in addition to the desired output, a trend also noted by others \citep{zhou2024larger}.}.

\subsection{Lexicon Diagnosis and Expansion}

\begin{figure}[t!]
\vskip 0.05in
\begin{center}
\includegraphics[width=5in]{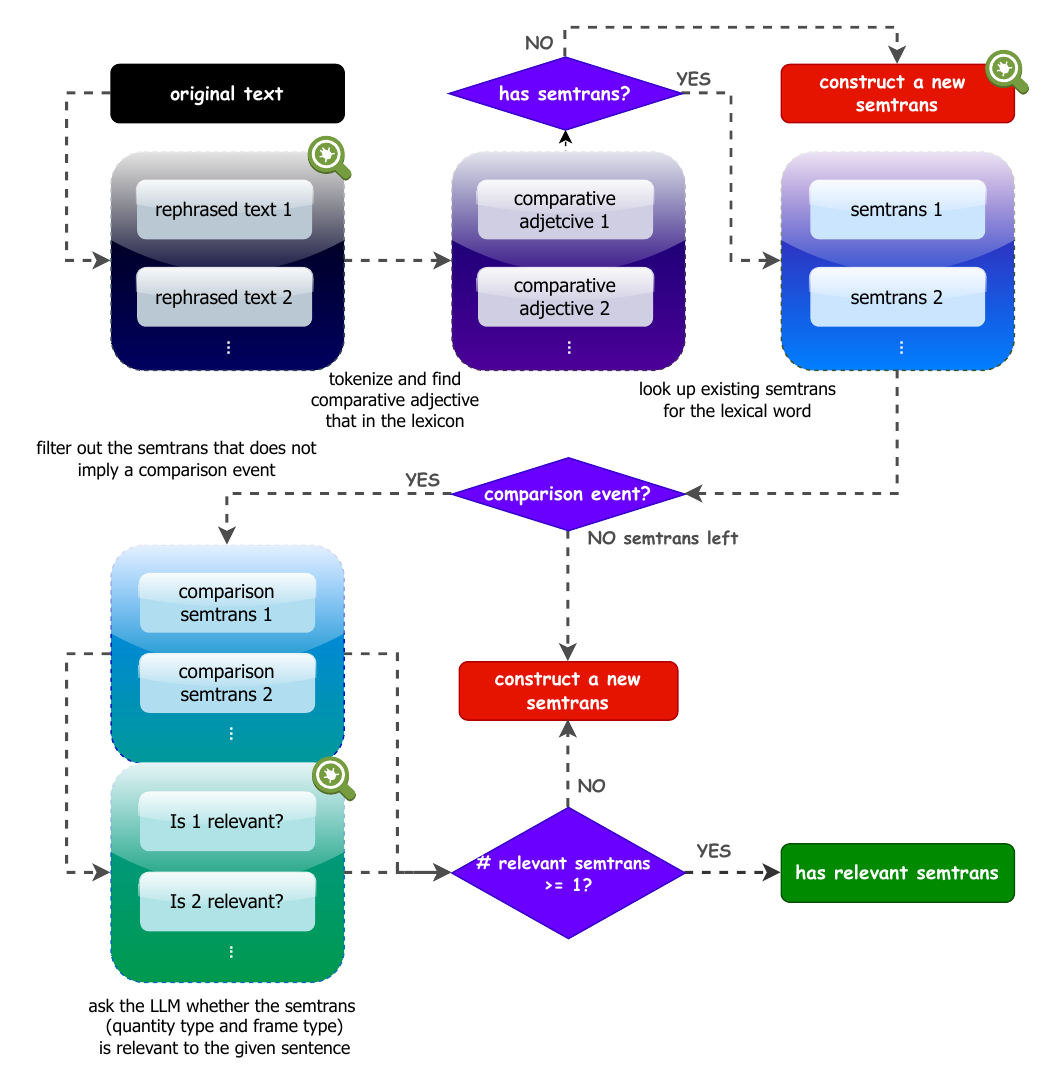}
\caption{Procedure for diagnosing an existing semtrans for a comparative adjective, together with its associated grounding fact under inspection. The diagnosis results in either (1) identifying at least one relevant semtrans from the existing knowledge base, or (2) constructing a new semtrans if: no semtrans exists for the adjective, no existing semtrans encodes a comparison event, or no existing semtrans is sufficiently relevant to the adjective as used in the given text. The dark green magnifying glass icon in the upper-right corner of a step indicates that an LLM assists the step.} 
\label{fig:semtrans-diagnose}
\end{center}
\vskip -0.2in
\end{figure}

Figure~\ref{fig:semtrans-diagnose} illustrates the procedure for checking and expanding the lexicon. The process is incremental and sequential—it processes one comparative adjective at a time rather than in batches, so each new evaluation consults the KB as updated by any semtrans added in earlier steps. In the following, we describe each step in detail: rephrased text generation, semtrans diagnosis, new semtrans construction, and determining frame-quantity relevance.

\subsubsection{Rephrased Texts Generation}

Given a grounding fact, we first prompt an LLM to generate rephrased sentence(s) in a fixed comparative structure designed to be easier for CNLU to parse and interpret: 
\begin{quote}
    {\scriptsize \tt [Noun] that is/has [comparative adjective] / [more/less quantity-noun] is/has \newline
    [comparative adjective] / [more/less quantity noun] than [Noun].}
\end{quote}
In this way, given the original grounding fact example ``If the gas is cooled, the particles will move more slowly, because they will have less energy'', the LLM produces:
\begin{quote}
    {\it Particles that are in cooler gas move more slowly than particles.}\\
    {\it Particles that are in cooler gas have less energy than particles.}
\end{quote}

\subsubsection{Semtrans Diagnosis}

For each rephrased text derived from the original grounding fact, CNLU first tokenizes the sentence and identifies comparative adjectives using the lexicon. For each comparative adjective, CNLU then searches the KB for any existing semtrans entries. If none are found, the system proceeds directly to a new semtrans generation. Otherwise, it filters the retrieved semtrans to retain only those that imply a comparison event (see Section~\ref{sec:CNLU} for an example). If no semtrans remain after this filtering, the system likewise proceeds to new semtrans generation.

When one or more semtrans introduce a comparison event for the comparative adjective, the system asks the LLM to assess whether each semtrans is sufficiently relevant to the given grounding fact, with particular attention to the quantity type and frame type it specifies. If none of the semtranses are relevant, the system proceeds to construct new semtranses.

\subsubsection{New Semtrans Construction}

\begin{figure}[t!]
\vskip 0.05in
\begin{center}
\includegraphics[width=6in]{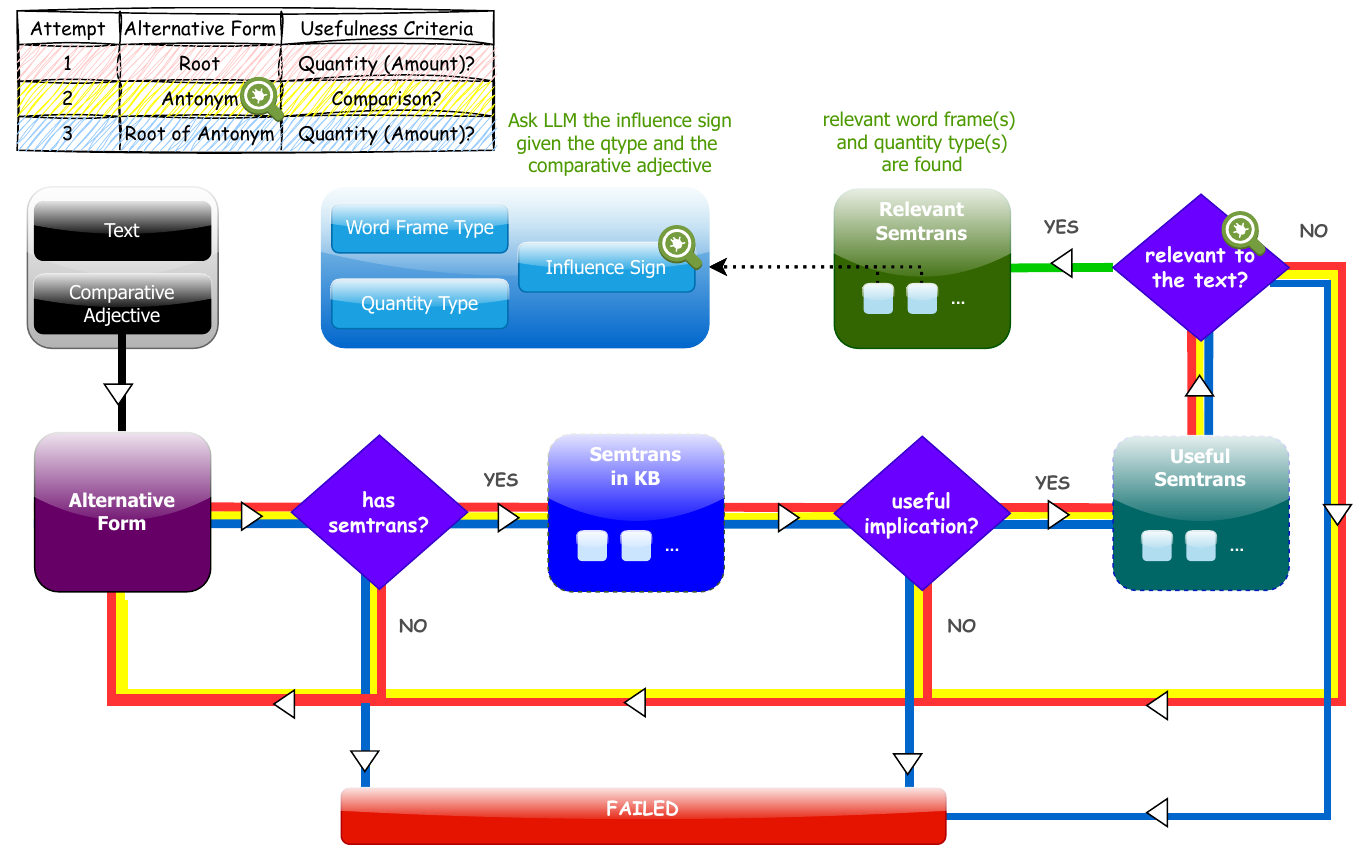}
\caption{Procedure for constructing a new semtrans for a comparative adjective deemed insufficiently relevant to the given grounding fact. The system begins by searching the relevant semtranses of the adjective's root form (red path), followed by the relevant semtranses of its antonyms (yellow path), and finally the relevant semtranses of each antonym's root form (blue path). If any search yields semtranses relevant to the grounding fact, the procedure stops and returns the identified quantity type(s) and frame type(s), along with the corresponding influence sign(s) obtained via the LLM. Using these results, the system generates new semtrans(es). If no relevant semtranses are found, the procedure terminates without generating new ones. A dark green magnifying glass icon in the upper-right of any step indicates LLM assistance.}
\label{fig:new-semtrans}
\end{center}
\vskip -0.2in
\end{figure}

Figure~\ref{fig:new-semtrans} outlines the process for constructing a new semtrans for a comparative adjective deemed insufficiently relevant to the grounding fact. The system first queries the lexicon for the adjective’s root form (e.g., {\it dense} for {\it denser}) and runs a similar diagnostic process: (1) If the root form has semtrans in the KB, continue; (2) Retain the semtrans of the root form that includes a qualitative value of a particular type of quantity (e.g. {\scriptsize \tt (HighAmountFn Density)} for {\it dense}); (3) Look for the quantity type given the qualitative value (e.g. {\scriptsize \tt Density} for {\it dense}); (4) Ask the LLM whether the quantity type is relevant given the comparative adjective's use in the original grounding fact; (5) Adopt the word frame and quantity type in the semtrans that is relevant. So as an example, given the comparative adjective {\it denser}, the system retrieves its root form {\it dense} and uses its relevant semtrans to collect the word frame and quantity types ({\scriptsize \tt FN\_Misc} and {\scriptsize \tt Density}). 



If it fails to retrieve any relevant semtrans from the root form, the system turns to its antonym (obtained via LLM, as the lexicon does not store antonym information) and applies the same diagnostic process. For instance, to construct a semtrans for {\it closer}, assuming its root form {\it close} does not have any relevant semtrans given some grounding fact, the system then inspects its antonym {\it farther} and adopts the most relevant word frame–quantity type pair ({\scriptsize \tt FN\_Gradable\_proximity} and {\scriptsize \tt Distance}) after evaluation.



Finally, suppose it still fails to find any relevant semtrans from the antonym form. In that case, it queries the root form of the antonym from the lexicon, then follows the same process as before to see whether a relevant word-quantity-frame pair exists in the KB. For instance, if it cannot find any relevant word-quantity-frame pair for original comparative adjective {\it closer} after inspecting its root form {\it close} and antonym {\it further}, it then inspects the root form of its antonym {\it far}. 

Once the appropriate word frame and quantity types are identified -- whether from the root form, the antonym, or the root form of the antonym -- the system queries the LLM for the influence sign, given the quantity type and comparative adjective. With the word frame type, quantity type, and influence sign determined, the new semtrans is created and added to the KB, making it available for future semtrans diagnoses.

\subsubsection{Determining Frame-Quantity Relevance}

Given a semtrans, specifically its word frame and quantity type, the system queries the LLM to determine whether they are relevant to a given grounding fact, taking into account the original use of the comparative adjective in that fact. For example, suppose the grounding fact is ``If the current passing through the circuit breaker increases, the electromagnet becomes stronger'', where the comparative adjective is {\it stronger}, in the case when its root form ({\it strong}), its antonym (e.g., {\it weaker}), or the root form of its antonym (e.g. {\it weak}) is inspected. The system has access to candidate frame-quantity-type pairs such as {\scriptsize \tt (Fn\_Usefulness Effectiveness)}, {\scriptsize \tt (FN\_Expertise Expertise)}, and {\scriptsize \tt (FN\_Level\_of\_force\_exertion Strength)}\footnote{In CNLU, these frame or quantity types, which are raw entities in the KB, are converted to natural language by a built-in natural language generation system to make sure the LLM understands what the entities mean. For example, given frame type {\tt FN\_Level\_of\_force\_exertion}, the query to the LLM would just mention it as ``level of force exertion''.}. The LLM is then asked, for each pair, whether it is relevant to the comparative adjective in the grounding act. In this case, the first two pairs should be judged irrelevant, while the last pair is relevant.
Note that one alternative way to determine the relevance between a frame-quantity-type pair and the grounding fact is to compute a ``relevance score'' quantitatively. This could be done by combining semantic similarity, comparative cues, and part-of-speech-based comparative patterns, and then evaluating whether an auto-generated sentence based on the frame-quantity-type pair lies sufficiently close to the grounding fact in the sentence embedding space \citep{reimers2019sentence, ethayarajh2019contextual}. However, this approach is less reliable: semantic relevance may not be fully captured by proximity in the embedding space alone, and our preliminary experiments showed that its performance was markedly lower than that of determining relevance by directly querying the LLM.

\subsection{Pipelines}

In our experiments, we evaluated the following pipelines:
\begin{itemize}
    \item {\bf Pure LLM}: We used Phi-4 to generate the desired logical forms directly from curated prompts containing a few illustrative examples.
    \item {\bf Pure CNLU}: We used CNLU only -- it does not use either the updated lexicon or the LLM-based sentence rephrasing. It generates semantic interpretations directly from the original text, produces QP frames based on these interpretations with narrative functions \citep{forbus2020analogies}, and then the QP frames are converted into readable assertions from which the quantity types and their influence signs can be extracted.
    \item {\bf Hybrid}: The hybrid pipelines generate the desired logical forms by CNLU with the assistance from Phi-4 -- so CNLU with either the updated lexicon, the sentence rephrasing with Phi-4, or both.
\end{itemize}
The pure CNLU and hybrid pipelines are expected to produce logical forms such as:
\begin{lstlisting}
(``When particles move more slowly, temperature is lower and an object feels cooler.'' (Speed -) (Temperature -))
\end{lstlisting}

\noindent Here, each output includes the original grounding fact along with quantity–sign pairs. In contrast, the pure LLM pipeline is expected to produce logical forms in the following format:
\begin{lstlisting}
(``When particles move more slowly, temperature is lower and an object feels cooler.'' (cause speed) (effect temperature) (cause-sign -) (effect-sign -))
\end{lstlisting}

\noindent Note that, when there is more than one cause or effect quantity type, the number of signs should match the number of quantity types in the logical form, like the example in the following:
\begin{lstlisting}
(``Bigger stars produce more energy, so their surfaces are hotter.''
 ((cause size) (effect amount-energy temperature) (cause-sign +) (effect-sign + +))
\end{lstlisting}


\subsection{Evaluation Metrics}
For a tested grounding fact $i$, let the gold quantity-sign specification be $Q_i$ and $s_i: Q_i \rightarrow \mathcal{S}$, where $Q_i$ is the set of gold quantity types and $\mathcal{S} = \{+, -, None\}$\footnote{If None, the system fails to find the sign for the found quantity type.} the set of influence signs. Each evaluated pipeline outputs the set of quantity types $\hat{Q_i}$ with their respective influence signs $\hat{s_i}: \hat{Q_i} \rightarrow \mathcal{S}$. 

To allow partial credit for near quantity type matches, while never rewarding signs when the quantity type is wrong, we define for each gold $q\in Q_i$, a matched prediction $\hat{q}\in \hat{Q_i} \cup \varnothing$, and a quantity match weight (in the experiments we set $\alpha=0.5$):
\[
w_i(q) = 
\begin{cases}
    1, & \text{exact, or ontology-equivalent match} \\
    \alpha \in (0, 1), & \text{partially relevant} \\
    0, & \text{not match}
\end{cases}
\]

Then the pipeline will have the following metrics regarding the tested grounding fact:
\begin{itemize}
    \item {\bf Quantity Coverage (QC) Score}: measures how well the system recovers the intended quantity types (full or partial matches), given by
    \[
    \frac{1}{|Q_i|}\sum_{q\in Q_i}w_i(q)
    \]
    \item {\bf Conditional Sign Accuracy (CSA) Score}: defined only when $\sum_{q\in Q_i}w_i(q) > 0$, otherwise reported as N/A; so if a quantity type is not matched, its sign still contributes no credit. It is given by
    \[
    \frac{\sum_{q\in Q_i} w_i(q) \cdot \mathbf{1}[s_i(q) = \hat{s_i}(\hat{q})]}{\sum_{q\in Q_i} w_i(q)}
    \]
    \item {\bf Overall Pair (OP) Score}: yields a single scalar per fact that (1) rewards finding the appropriate quantity types (with partial credit possible), and (2) rewards getting the sign correct only when the quantity is (at least partially) correct. It is given by
    \[
    \frac{\sum_{q\in Q_i} w_i(q) + \sum_{q\in Q_i} w_i(q) \cdot \mathbf{1}[s_i(q) = \hat{s_i}(\hat{q})]}{2|Q_i|}
    \]
\end{itemize}

For example, consider the grounding fact ``When a gas is squeezed into a smaller volume, the particles have less space to move'' whose gold quantity–sign pairs are {\scriptsize \tt (Volume -) (SpatialQuantity -)}. Suppose the system predicts {\scriptsize \tt (Volume -) (Size -)}. By assigning a partial quantity weight of 0.5 to {\scriptsize \tt Size} as a near match to {\scriptsize \tt SpatialQuantity} and a quantity weight of 1.0 to {\scriptsize \tt Volume} as an exact match,  the QC score becomes \[\frac{1.0 + 0.5}{2} = 0.75\]
Since the returned signs for both quantity types are accurate, the CSA score is given by \[\frac{1.0\cdot 1 + 0.5\cdot 1}{1.0 + 0.5} = 1.0\]
Finally, the OP score is given by \[\frac{(1.0 + 0.5) + (1.0\cdot 1 + 0.5\cdot 1)}{2\cdot 2}=0.75\]
indicating an overall ``matching'' score. Note that, these metrics assess coverage of the gold quantity-sign pairs rather than the exact equality of the gold and generated quantity-sign pairs. After computing scores for all tested grounding facts, we report averages for each evaluated pipeline, each times 100\% as the actual score out of 100; for conditional sign accuracy, we average only over facts with at least one matched quantity (i.e., nonzero denominator).

\section{Results}

By using 282 distinct grounding facts in the QuaRTz training set, the system constructed 28 new semtrans facts, all of which are reasonable. After that, all pipelines generated the expected logical form, indicating the associated quantity types and respective influence signs, given the 81 distinct grounding facts in the QuaRTz test set. Their metric scores are shown in Table~\ref{tab:metrics}.

\begin{table}[t!]
\vskip -0.15in
\caption{Metric scores for each pipeline.}
\label{tab:metrics}
\small
\begin{center}
\begin{tabular}{c|c||c|c|c}
\hline\hline
\multicolumn{2}{c||}{{\bf Pipeline}} & {\bf QC} & {\bf CSA} & {\bf OP} \\
\hline\hline
\multicolumn{2}{c||}{CNLU} & 16.05 & 90.00 & 11.78 \\
\hline
\multirow{3}{3em}{Hybrid} & Updated Lexicon & 22.33 & 91.05 & 17.44 \\
& Text Rephrase & 26.60 & 94.44 & 22.62 \\
& Both & 34.67 & 90.76 & 30.90 \\
\hline
\multicolumn{2}{c||}{Phi-4} & 89.86 & 95.21 & 87.88 \\
\hline
\end{tabular}
\end{center}
\vskip -0.10in
\end{table}

\subsection{LLM-Only}

The Phi-4–only pipeline performs strongly on extracting quantity types and their influence signs. In most test cases, its generated logical forms cover the required quantities implied by the grounding facts and assign the correct influence signs to those quantities, yielding high QC and CSA scores. It achieves an OP score of 87.88 (0–100 scale), outperforming the other pipelines on our evaluation.

However, it still fails in some cases that humans find straightforward, and CNLU or hybrid pipelines can solve better. For example, given the test case ``Convection on early Earth was faster and so plate tectonics was faster,'' the relevant quantity types should be the Earth's time or age (cause) and the event rate of plate-tectonic motion (effect). Phi-4 instead returned {\scriptsize \tt convection} and {\scriptsize \tt speed-plate-tectonics}. The former receives no credit because ``convection'' is a process or entity, and even if it can be wrapped as a quantity type, it would still be irrelevant to the time or age of the Earth; the causal quantity must be the \emph{age of the Earth}. The latter is also not perfect: ``speed'' typically denotes motion over distance per unit time, whereas the intended construct is an \emph{event rate} (e.g., {\scriptsize \tt EventRate}); this is captured more appropriately by the CNLU or hybrid pipelines. 
More broadly, this kind of error is consistent with an association-based inference strategy: an LLM tends to surface distributionally ``relevant'' lexical items (e.g. {\it convection}, {\it speed}) that co-occur with cues like {\it faster}, rather than projecting the text into the KB's quantity ontology and enforcing causal role constraints. In effect, it retrieves salient entities instead of the latent quantities that license the inference (here {\scriptsize \tt Age} or {\scriptsize \tt AGE(Earth)} as cause, and {\scriptsize \tt EventRate} as effect), which has been illustrated as the shortcut or heuristic behavior in language models \citep{bender2020climbing, mccoy2019right}. This also aligns with our metric pattern -- high CSA conditional on a matched quantity, but lower QC when entity-quantity disambiguation is required, suggesting probabilistic association rather than quantity-level reasoning in such cases.

The LLM can also introduce occasional structural errors in the logical forms, such as missing signs or fusing multiple quantities into a single term. For instance, the following output omits the effect sign for {\scriptsize \tt depth}:
\begin{lstlisting}
("Older layers are deeper in the Earth, younger layers are closer to the surface."
  ((cause age) (effect depth) (cause-sign +)))
\end{lstlisting}
In another case, multiple distinct quantity types are merged into an integrated label, leading to inconsistent formal representations:
\begin{lstlisting}
("A gentle slope favours slower flow of surface water, reduces erosion, and
 increases availability of water to plants."
  ((cause slope) (effect speed-erosion-availability-water)
   (cause-sign -) (effect-sign - - +)))
\end{lstlisting}

\noindent Beyond such formatting errors, two practical concerns remain:
\begin{itemize}
  \item {\bf KB interoperability.} Terms in the generated logical forms often lack a direct mapping to entities in the KB, limiting immediate consumption by symbolic components such as CNLU.
  \item {\bf Potential data contamination.} QuaRTz was designed to inspect neural models’ reasoning abilities, and has been available on the open web since its publication. Hence it is very likely that the entire dataset has been used as part of the training corpus of LLMs. System performance may therefore depend on prior exposure rather than general language or reasoning capabilities.
\end{itemize}
Despite these issues, within our setup, the LLM-only pipeline is generally much more reliable at identifying relevant quantity types than the CNLU-only and hybrid variants.

\subsection{CNLU-Only and Hybrid}

On longer and syntactically complex grounding facts, the CNLU-only pipeline frequently fails to produce full parses or reliable semantic choices, and—even when parses are available -- often misses quantity types implied by the sentence. Nevertheless, it attains a QC score of 16.05 and, despite relatively strong sign handling on the quantities it does find (high CSA score), its overall score remains modest with the OP score 11.78, reflecting low coverage.

As mentioned, augmenting CNLU with Phi-4 improves performance via two mechanisms:
\textbf{(i) Lexicon update.} With Phi-4 assisting lexicon expansion (while retaining CNLU’s original syntactic parsing), the pipeline reaches a QC score of 22.33 and an OP score of 17.44.
\textbf{(ii) Text rephrasing.} When Phi-4 rephrases grounding facts into more CNLU-amenable forms, the pipeline attains the QC score 26.60 and a high CSA score 94.44, yielding the OP score 22.62.
\textbf{(iii) Both lexicon update and rephrasing.} Combining both interventions yields the best hybrid results: A QC score of 34.67 and an OP score of 30.90. In other words, the hybrid pipeline correctly covers more than one-third of the required quantity types, while remaining imperfect on some signs.

Still, there are some limitations of the hybrid approach that hinder it from having comparable performance with Phi-4:
\begin{itemize}
  \item \textbf{Incomplete coverage of comparative determiners.} The current lexicon diagnosis and expansion targets comparative adjectives that directly signal comparisons over their implied quantity types. Degree determiners and noun-modifying comparatives (e.g., \emph{more}, \emph{less}, \emph{higher \textlangle quantity\textrangle}, \emph{lower \textlangle quantity\textrangle}) are not yet handled well in context, and such cases are common in QuaRTz and central to describing quantity change.
  \item \textbf{Residual complexity after rephrasing.} Some LLM rephrasings remain multi-clausal or otherwise difficult for CNLU to parse or to map to QP frames. For example, rephrasing ``Friction causes the molecules on rubbing surfaces to move faster, so they have more energy. This gives them a higher temperature, and they feel warmer.'' as ``Surfaces that are rubbed together have molecules with more energy than surfaces that are not rubbed together.'' still leaves structures that challenge CNLU’s narrative functions.
  \item \textbf{Occasional irrelevant rephrasings.} Despite our having tried to avoid generations of such texts from Phi-4, it sometimes still produces meta-level or explanatory text that is off-task, or plausible-sounding rephrasings that are semantically unrelated to the original fact, partially due to its hallucinations.
  \item \textbf{Over-rephrasing of simple inputs.} Not all grounding facts should be rephrased. For instance, ``More people need more resources'' is already canonical for CNLU; rephrasing it to ``People that are more in number need more resources than people'' distorts the original meaning and harms downstream analysis.
\end{itemize}

\noindent Taken together, these results show that LLM assistance can substantially improve CNLU’s ability to identify quantities and their influence signs from raw grounding facts. With better coverage of comparative determiners, stricter controls on rephrasing, and continued lexicon curation, the hybrid approach has clear headroom to approach LLM-only performance while offering greater interpretability and controllability, and without requiring extensive task-specific training data.

\section{Conclusions and Future Work}

In this paper, we presented our efforts to integrate an LLM with the symbolic NLU system CNLU for generating desired logical forms and interpreting continuous quantity changes from continuous causal statements. In our approach, LLMs are used to rephrase original texts into a regularized format that is easier for CNLU to parse, and to play targeted roles in updating the semantic knowledge base. These roles include supplying lexical information (e.g., antonyms) and semantic information (e.g., the influence sign of a quantity implied by a comparative adjective), as well as supporting statistical methods that use embedding spaces and external lexicons to determine whether a quantity is semantically related to the given text. Our results show that this hybrid method significantly outperforms a purely symbolic NLU system, providing broader language coverage while preserving the advantages of inspectability, support for high-precision reasoning, and incremental rapid learning that symbolic systems provide.

This paper is an initial exploration of incorporating LLMs into the interpretation of natural language texts within a specific reasoning domain (qualitative reasoning) using a massive knowledge base with complex, hand-curated ontologies. Potential future work spans several directions. First, to better handle comparative adjectives that modify the degree or amount of a noun or quantity (e.g., ``more water'', ``less acceleration''), we need encoding strategies beyond semtrans supplementation, for example, more comprehensive grammar rules that directly capture such expressions. Second, the semtrans update procedure should be expanded to incorporate richer methods for identifying potential quantity types. Third, rephrased sentence generation needs to be made more robust. One possible approach is to have LLMs generate structured, fine-grained sentence elements, which CNLU could then assemble into complete sentences according to defined rules. Finally, once the hybrid NLU approach achieves more reliable performance in generating relevant interpretations of quantities, the targeted quantities diagnosed during the semtrans process could be used to specify the quantity types that the NLU system should map to. Possible ways include analogical Q/A training, using query cases constructed from rephrased texts paired with their logical forms, including target quantities only, or using a variation-set like method to invoke an LLM to generate more semantically similar but syntatically different sentences mapping to the logical forms we want.



\begin{acknowledgements} 
\noindent
This research was supported by the Office of Naval Research. Additionally, we extend our sincere appreciation to individuals at Qualitative Reasoning Group who generously shared their invaluable insights and feedback on this work, notably Constantine Nakos,  Wangcheng Xu, Zoie Zhao, Omar Khater, and Jiahong Zheng.
\end{acknowledgements} 


{\parindent -10pt\leftskip 10pt\noindent
\bibliographystyle{cogsysapa}
\bibliography{format}

}


\end{document}